\documentclass[sigconf,natbib=true]{acmart} %
\usepackage{bbm}
\usepackage{multirow}
\usepackage{listings}
\usepackage{xcolor}
\usepackage{csquotes}
\usepackage{blindtext}

\renewenvironment{quote}
  {\small\list{}{\rightmargin=0.3cm \leftmargin=0.3cm}%
   \item\relax}
  {\endlist}

\definecolor{codegreen}{rgb}{0,0.6,0}
\definecolor{codegray}{rgb}{0.5,0.5,0.5}
\definecolor{codepurple}{rgb}{0.58,0,0.82}
\definecolor{backcolour}{rgb}{0.95,0.95,0.92}

\lstdefinestyle{mystyle}{
	backgroundcolor=\color{backcolour},   
	commentstyle=\color{codegreen},
	keywordstyle=\color{magenta},
	numberstyle=\tiny\color{codegray},
	stringstyle=\color{codepurple},
	basicstyle=\ttfamily\footnotesize,
	breakatwhitespace=false,         
	breaklines=true,                 
	captionpos=b,                    
	keepspaces=true,                 
	numbers=left,                    
	numbersep=5pt,                  
	showspaces=false,                
	showstringspaces=false,
	showtabs=false,                  
	tabsize=2,
	escapeinside={(*}{*)}
}

\newcommand{\squishlist}{
	\begin{list}{$ullet$}
		{ \setlength{\itemsep}{0pt}
			\setlength{\parsep}{3pt}
			\setlength{\topsep}{3pt}
			\setlength{\partopsep}{0pt}
			\setlength{\leftmargin}{1.5em}
			\setlength{\labelwidth}{1em}
			\setlength{\labelsep}{0.5em} } }
	
	\newcommand{\squishend}{
\end{list}  }

\usepackage{lipsum}

\AtBeginDocument{%
  \providecommand\BibTeX{{%
    \normalfont B\kern-0.5em{\scshape i\kern-0.25em b}\kern-0.8em\TeX}}}

\copyrightyear{2023}
\acmYear{2023}
\acmConference[Gen-IR@SIGIR2023]{The First Workshop on Generative Information Retrieval}{July 27, 2023}{Taipei, Taiwan}
\acmBooktitle{The First Workshop on Generative Information Retrieval (Gen-IR@SIGIR23), July 27, 2023, Taipei, Taiwan}
\acmDOI{}
\acmPrice{}
\acmISBN{}
\setcopyright{rightsretained}

\begin{document}

\title{On Exploring the Reasoning Capability of Large Language Models with Knowledge Graphs}

\author{Pei-Chi Lo }
\email{pclo.2017@phdcs.smu.edu.sg}
\affiliation{%
  \institution{Singapore Management University}
  \country{Singapore}
}

\author{Yi-Hang Tsai}
\email{d094020002@nsysu.edu.tw}
\affiliation{%
  \institution{National Sun Yat-sen University}
  \country{Taiwan}
}
\author{Ee-Peng Lim}
\email{eplim@smu.edu.sg}
\affiliation{%
  \institution{Singapore Management University}
  \country{Singapore}
}

\author{San-Yih Hwang}
\email{syhwang@mis.nsysu.edu.tw}
\affiliation{%
  \institution{National Sun Yat-sen University}
  \country{Taiwan}
}

\begin{abstract}
This paper examines the capacity of LLMs to reason with knowledge graphs using their internal knowledge graph, i.e., the knowledge graph they learned during pre-training. Two research questions are formulated to investigate the accuracy of LLMs in recalling information from pre-training knowledge graphs and their ability to infer knowledge graph relations from context. To address these questions, we employ LLMs to perform four distinct knowledge graph reasoning tasks. Furthermore, we identify two types of hallucinations that may occur during knowledge reasoning with LLMs: content and ontology hallucination. Our experimental results demonstrate that LLMs can successfully tackle both simple and complex knowledge graph reasoning tasks from their own memory, as well as infer from input context.
\end{abstract}

\maketitle

%% your paper content here
%%%%%%%%%%%%%%%%%%%%%%%%%%%%%%%%%%%%

\section{Introduction}
	\label{sec:intro}
	Large Language Models (LLMs), with the ability to in-context learning and Chain-of-Thought (CoT), have shown to outperform previous state-of-the-art models in many information retrieval (IR) tasks~\cite{Wei_Wang_Schuurmans_Bosma_Ichter_Xia_Chi_Le_Zhou_2022,Wei_Tay_Bommasani_Raffel_2022,Chen_2022,Chen_Zhang_Xie_Deng_Yao_Tan_Huang_Si_Chen_2022,Dai_Sun_Dong_Hao_Sui_Wei_2022,Dong_Li_Dai_Zheng_Wu_Chang_Sun_Xu_Li_Sui_2022,Huang_Chang_2022,Nye_Andreassen_Gur_2022}, e.g., question answering~\cite{Chen_Ma_Wang_Cohen_2022,Chen_Jia_Ding_Shen_Xiang_2020,Creswell_Shanahan_Higgins_2022,Huang_Gu_Hou_Wu_Wang_Yu_Han_2022,Katz_Geva_Berant_2022} and common sense reasoning~\cite{Liu_Liu_Lu_Welleck_West_2022,Madaan_Zhou_Alon_Yang_Neubig_2022}. 
	In this work, we explore LLMs' ability in addressing knowledge graph reasoning with its own knowledge capacity. 
	
	To the best of our knowledge, no examination has been conducted on the ability of LLMs to deduce knowledge from its internal Knowledge Graphs (KGs). As the training process and training data are both undisclosed for mainstream LLMs such as \textsf{text-davinci-003} and \textsf{ChatGPT}, it is unknown whether LLMs are pre-training on knowledge graph data. In this study, we therefore investigate the capacity of LLMs to recall information from their internal KGs, i.e., the knowledge graph they learned during pre-training. Specifically, we examine their ability to deduce direct relations, which refer to one-hop connections in the KG. We then proceed to explore their capability to infer multi-hop relations, which is a more challenging task that requires not only relation inference but also the formation of paths in the KG. In the area of information retrieval, contextual information plays a crucial role in disambiguating words that have multiple meanings. As such, we investigate whether LLMs have the ability to retrieve contextual information from a given text.
	
	In this study, our first focus is to investigate the capacity of LLMs to retrieve information related to knowledge graphs from their own memory. To address this issue, our first research question is formulated as follows: \textbf{RQ1: To what extent can LLMs accurately recall information from KG?}
	To address this research question, we employ LLMs to perform two non-contextual relation generation tasks, namely \textbf{tail entity prediction} and \textbf{relation prediction}.
	
	During the experiment, we observe how LLMs suffer from \textit{hallucination} when addressing these tasks. This raises concerns of deploying LLMs in practice and compromises result accuracy as hallucinations result in nonfactual statements and can potentially be harmful ~\cite{Dong_Li_Dai_Zheng_Wu_Chang_Sun_Xu_Li_Sui_2022,Huang_Chang_2022}. We define hallucination in knowledge graph reasoning into two types: \textit{content} and \textit{ontology hallucination}.  Previous works address the hallucination issue by introducing an additional retrieval step that retrieves external information such as Wikipedia pages ~\cite{Bohnet_Tran_Verga_2022,He_Zhang_Roth_2022,Trivedi_Balasubramanian_Khot_Sabharwal_2022} and condition the LLM's generation on both the query input and retrieved external information.  This results in a more controlled result generation with less hallucination.

	% ability to retrieve information from context
	An essential aspect of reasoning with knowledge graphs is the capacity to deduce relations within the graph based on contextual information. This ability is critical, and it leads us to our second research question: \textbf{RQ2: To what extent can LLMs infer knowledge graph relations from context?} To address this question, we undertake two tasks that involve generating contextual relations using LLMs: Relation Extraction (RE) and Contextual Path Generation (CPG).

	In the following sections, we present our contributions as follows: (1) we investigate two research questions that explore the LLMs' ability to reason with their internal knowledge graph, (2) we distinguish content and ontology hallucination that occur in knowledge reasoning tasks, and (3) experiment results indicate that the LLMs can retrieve knowledge graph information from memory and infer knowledge graph relations from given context.

	\section{Recalling Knowledge Graph Information within LLM}
		%In this section, we approach two non-contextual relation generation tasks with LLMs to address our first research question: How well can LLMs retrieve the knowledge graph information within itself?
		
		\subsection{Non-contextual Relation Generation}\label{sec:nc_rel_gen}
		The first objective of our study is to investigate whether LLMs have been exposed to knowledge graph data during pre-training, and their ability to recall such information. Here, we focus on two knowledge graph relation generation tasks, namely \textbf{tail entity prediction} and \textbf{relation prediction}. 
		The former task involves identifying the tail entity based on the input <\textsf{head entity, \textit{relation}, ?}>. The latter task requires the model to recover the relation based on the input <\textsf{head entity, ?, tail entity}>. 
		
		As no context is provided during the generation process, both of these tasks requires the LLMs to to remember and utilize the knowledge graph structure learned during pre-training. In other words, high non-contextual relation generation accuracy indicates that (1) LLMs have seen knowledge graph data during pre-training, and (2) they are able to retrieve from this internal knowledge graph.
		
		\subsubsection{Dataset}
		We selected a random sample of 100 <head entity, relation, tail entity> triples from DBpedia. To create the tail entity prediction dataset, we generated queries by masking the tail entity from each triple. Similarly, we constructed the relation prediction dataset by masking the relation from the same set of 100 triples. It is important to note that some queries may have multiple correct answers for tail entity or relation prediction tasks. 
		
		For example, for the query <\textsf{dbr:Moneyball\_(film) - \textit{dbo:starring} - ?}>, both \textsf{dbr:Brad\_Pitt} and \textsf{dbr:Jonah\_Hill} are considered valid answers. Likewise, \textsf{\textit{dbo:producer}} and \textsf{\textit{dbo:starring}} are both consider correct for the query <\textsf{dbr:Moneyball\_(film) - ? - dbr:Brad\_Pitt}>

		\subsubsection{Evaluation Metrics}
		In this study, we present two evaluation metrics, namely \textbf{Hard Accuracy (H-ACC)} and \textbf{Soft Accuracy (S-ACC)}. A generation is considered hard accurate if it meets both of the following criteria: (1) the outputted <head entity, relation, tail entity> triple exists relation in DBPedia, and (2) it has the same relation/entity surface as the one presented in DBPedia. 
		On the other hand, soft accuracy is a more lenient evaluation metric. It considers a generation to be accurate as long as the relation/entity generated is factually correct, regardless of whether the surface of the relation/entity is identical to that in DBPedia. 
		
		To illustrate, given the query \textsf{dbr:William\_Thomas\_Hamilton} - ? -   \textsf{dbr:Boonsboro,\_Maryland}, we only consider the ground truth \textsf{\textit{dbo:birthPlace}} to be hard-accurate. However, generations such as \textsf{\textit{dbo:bornIn}} and \textsf{\textit{dbo:birthLocation}} will be consider as soft-accurate because they are synonymous to the ground truth. 
		We manually label the hard and soft accuracy and report the average performance metrics based on 10 generation trials.
		
		\subsubsection{Models}
		In this study, we employed three text-based large language models, namely \textsf{text-davinci-003}, \textsf{ChatGPT}, and \textsf{GPT-4}. For \textsf{text-davinci-003} and \textsf{ChatGPT}, we set the temperature to 0. All of the experiments are conducted under zero-shot setting.
		
		We employed the same prompt format across all models. For tail entity generation:
		\begin{quote}
			Complete the following DBPedia relation:\\
			\textbf{<head entity>} - \textbf{<relation> - }
		\end{quote}
		
		For relation generation:
		\begin{quote}
			Insert DBPedia ontology property between the following DBpedia entity pair to form an existed DBPedia relation:\\
			\textbf{<head entity>}, \textbf{<tail entity>}
		\end{quote}

		\begin{table}[t!]
			\small
			\caption{Non-contextual Relation Generation}
			\label{tbl:nc_rel_gen}
			\begin{tabular}{ccccc}
				\toprule
				\multirow{2}{*}{} & \multicolumn{2}{c}{\textbf{Tail Ent Generation}} & \multicolumn{2}{c}{\textbf{Relation Generation}} \\
				& \textbf{H-ACC} & \textbf{S-ACC} & \textbf{H-ACC} & \textbf{S-ACC} \\
				\midrule
				text-davinci-003 & 48.2 & 54.6 & 47.6 & 87.4 \\
				ChatGPT & 64.5 & 67.2 & 53.8 & 82.6 \\
				GPT-4 & 68.1 & 87.3 & 48.6 & 90.2\\
				\bottomrule
			\end{tabular}
        \vspace{-0.5cm}
		\end{table}

		\subsubsection{Result}
		We present the quantitative performance in Table~\ref{tbl:nc_rel_gen}, and selected case examples in Table~\ref{tbl:exp_nc_rel_gen}. In both types of non-contextual relation generation tasks, our results indicate that more advanced model result in improved relation generation accuracy. 
		
		Although the H-ACC between GPT-4 and ChatGPT seems comparable, GPT-4 produces significantly higher H-ACC. This suggests that GPT-4 generates relations that are more synonymous with the ground truth. For example, in the first example in Table~\ref{tbl:exp_nc_rel_gen}, GPT-4 produces dbr:Tomb\_Raider\_(soundtrack) - dbp:label - dbr:Milan\_Records, which is a record label owned by Sony Music and the ground truth. In contrast, ChatGPT generates a non-existent entity, Tomb Raider: Original Motion Picture Soundtrack. 
		
		It is worth noting that all of the models have generated factually correct result that are not ground truth, resulting in performance gap between H-ACC and S-ACC. For instance, in the second example shown in Table~\ref{tbl:exp_nc_rel_gen}, all models generate dbr:California\_Institute\_of\_the\_Arts for the query dbr:Tim\_Burton - dbo:education - ?, despite this entity-relation-entity triple not existing in DBPedia. Nevertheless, it is a fact that Tim Burton attended the California Institute of the Arts.
		
		Based on this outcome, it is sufficient to conclude that \textit{LLMs are capable of retrieving information from KG to a reasonable extent}. However, even the most advanced model produces a significant number of errors. In the subsequent section, we formally categorize two types of errors made by LLMs.

		\subsection{Hallucinations in LLM-based Knowledge Graph Reasoning}
		In our empirical evaluation of the DBPedia relations generated by LLMs, we found two major types of hallucination:\\
		(a) \textbf{content hallucination}, which refers to generated relations that do not exist or are non-factual.  For instance, path (a): \textsf{ dbr:Kate\_Winslet - {\color{red}\textit{ dbo:spouse}} - dbr:Jamie\_Foxx} is a non-factual relation.
		(b) \textbf{ontology hallucination}, which refers to generated path relations that are invalid according to DBPedia ontology. For instance, \textsf{dbr:Reading,\_Berkshire - {\color{red}\textit{dbo:location}} - dbr:Jamie\_Foxx} is invalid. Under DBPedia defintion, \textsf{\textit{dbr:location}} should only link to entities of type \underline{Location} while \textsf{dbr:Jamie\_Foxx} is of type \underline{Person}.
		
		In summary, \textit{hallucination occurs in knowledge graph reasoning by LLMs}. To help the large language models generate relations that are factual and fit to the ontology, we need to design prompts so as to reduce hallucination in the generated relations.

	\section{Inferring Knowledge Graph Relation from Context}
	\subsection{Contextual Relation Generation}	
	While LLMs have demonstrated their capability in retrieving information from within knowledge graphs, their ability to reason with this internal knowledge graph is still an open question. To address this research gap, we conducted a contextual relation generation task that required LLMs to comprehend a given context and employ the information from its internal knowledge graph to reason and generate relations.
	Relation Extraction (RE) task is a type of contextual relation generation task.
	Given a pair of entities and the context in which they are mentioned, the Relation Extraction task aims to predict a knowledge graph relation that connects the entity pairs from the context. In this paper, we approach the relation extraction task as a generation task. In other words, our goal is to use LLMs to generate a knowledge graph relation that completes a <head entity, ?, tail entity> triple based on the given context. 
	
	To successfully generate the missing relation, large language models needs to equip with the ability to (1) infer the semantic relationship between the query entities from the context and (2) generate an existing and valid knowledge graph relation that corresponds to the semantic relationship. 
	
	\subsubsection{Dataset}
		In this experiment, we augment the non-contextual relation generation dataset described in Section~\ref{sec:nc_rel_gen} to create a new dataset. The augmentation process involves selecting a context paragraph from the Wikipedia pages of the query entities for each query. To ensure the quality of the dataset, we conducted a manual examination, checking that both query entities were mentioned in the context paragraph, and that each query had only one ground truth relation.
		We present two example queries in Table~\ref{tbl:c_rel_gen}. In both examples, multiple DBPedia relations exist between the query entities. For instance, \textsf{\textit{dbo:founders}}, \textsf{\textit{dbo:owner}}, \textsf{\textit{dbo:keyPerson}} are all existed relations between \textsf{dbr:Playtone} and \textsf{dbr:Tom\_Hanks}. However, as the relation should be inferred from the context, only \textsf{\textit{dbo:founders}} should be considered a correct contextual relation generation.
		
	\subsubsection{Evaluation Metrics}
		As in Sec~\ref{sec:nc_rel_gen}, we present the average performance metrics of the \textbf{Hard Accuracy (H-ACC)} and \textbf{Soft Accuracy (S-ACC)} based on 10 generation trials. 

    		\begin{table}[t!]
			\small
			\caption{Contextual Relation Generation}
			\label{tbl:re}
			\begin{tabular}{ccc}
				\toprule
				& \textbf{H-ACC} & \textbf{S-ACC} \\
				\midrule
				text-davinci-003 & 63.2 & 76.4  \\
				ChatGPT & 72.8 & 78.4 \\
				GPT-4 & 74.2 & 83.5\\
				\bottomrule
			\end{tabular}
   \vspace{-0.3cm}
		\end{table}
 
	\subsubsection{Models}
		We employed the same three text-based large language models as in Sec~\ref{sec:nc_rel_gen}. For \textsf{text-davinci-003} and \textsf{ChatGPT}, we set the temperature to 0. All of the experiments are conducted under zero-shot setting.
		We employed the same prompt across all models:
		\begin{quote}
			Context: \textbf{<context>}
			
			Instruction: Insert DBPedia entities and ontology properties between \textbf{<head entity>} and \textbf{<tail entity>}  to form one single relation which reflects the relationship between them in the context. Pay attention to the relation format dbr:ENTITY\_NAME, dbo:ONTOLOGY, dbr:ENTITY\_NAME
		\end{quote}
		
	\subsubsection{Result}
		Table~\ref{tbl:re} presents the results of our experiment. Our investigation into non-contextual relation generation reveals that \textsf{GPT-4} outperforms all other models, achieving a H-ACC score of 74.2\%. \textsf{ChatGPT} follows closely with a H-ACC score of 72.8\%, while \textsf{text-davinci-003} performs the weakest among the three models, with a H-ACC score of only 63.2\%. This finding is consistent with our previous experiment, which suggests that more advanced LLMs lead to higher knowledge graph reasoning accuracy. As observed in Section~\ref{sec:nc_rel_gen}, we note a significant increase in performance from H-ACC to S-ACC. However, the performance gap is smaller compared to non-contextual relation generation. This may be due to the fact that H-ACC is already quite high, thereby limiting the potential for further performance improvement.
		
		We also observe that the models frequently err by producing the more popular relation. For instance, in the first example presented in Table~\ref{tbl:c_rel_gen}, ChatGPT generates \textsf{\textit{dbo:owner}} instead of the actual relation \textsf{\textit{dbo:founder}}. Similarly, in the second example, \textsf{text-davinci-003} generates \textsf{\textit{dbo:starring}} instead of the correct relation \textsf{\textit{dbo:producer}}. Both of these relations that were generated incorrectly are more commonly used (i.e., are more likely to connect to the query entities) than the actual relations. This highlights how the prior knowledge present within LLMs can still influence the generation of relations even when a context is given.
		According to the result, we can confidently suggest that \textit{LLMs are able to infer knowledge graph relation from context with a good level of accuracy}.

	\subsection{Contextual Path Generation}
\begin{figure}[t!]
			\centering
			\includegraphics[width=0.8\linewidth]{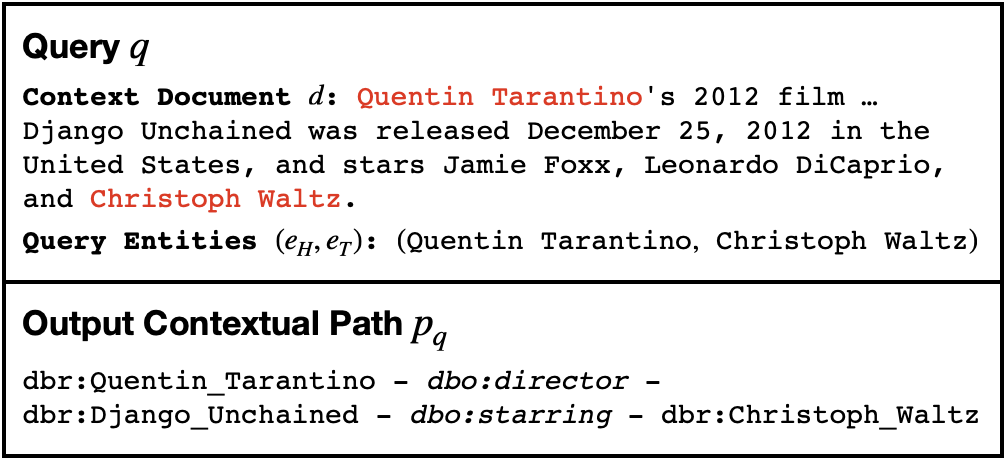}
			\caption{An Example of Contextual Path Generation Task. The query entities $(e_H, e_T)$ are colored in red in the context document $d$.}
			\label{fig:CPG}
			\vspace{-4mm}
		\end{figure}

	While Previous experiments have shown that LLMs are able to infer knowledge graph relations from context, we are interested in how LLMs perform in complex contextual knowledge graph reasoning tasks. In this experiment, we introduce the \textbf{Contextual Path Generation (CPG)} task which involves context-specific knowledge graph reasoning. CPG aims to infer a multi-hop knowledge graph path, called contextual path, to explain the semantic connection between two entities found in a context document. We show an example of CPG in Figure~\ref{fig:CPG} in which a query consists of a context document mentioning two entities Quentin Tarantino and Christoph Waltz. The semantic connection between the entities is a knowledge graph path that shows Waltz starred in the Django Unchained movie directed by Tarantino.	
	
	The existing LLM-based solutions usually generate reasoning steps in natural language text. Such unstructured output poses great challenge to not only downstream applications, but also performance evaluations. For complex knowledge graph reasoning tasks such as CPG, it is thus important to address the challenge of guiding LLMs to probe its memory of the relevant knowledge graph and to generate a well-formed and accurate path.  CPG also involves complex query input data which consist of context document and two query entities. We thus have to address the additional challenge of getting LLMs to encode these input data through prompting.

		\subsubsection{Dataset}
		We request the Contextual Path Retrieval (CPR) dataset from the authors of ~\cite{CPR}. The dataset contains 40 Wikinews articles as context documents, which mention 563 DBPedia entities. Due to budget and API limitation, we only sample 3 queries for each context document and derive a total of 120 CPG tasks. One ground truth contextual path is determined for each task. The hop count of contextual paths ranges from 2 to 6.

		\subsubsection{Evaluation Metrics}
		We evaluate the generated contextual paths by (a) path correctness and (b) path well-formedness.  For \textit{path correctness}, we measure the \textbf{Normalized Graph Edit Distance (NGEO)}~\cite{CPR}. NGEO of a generated sequence $s$ and the ground truth sequence $s^*$ can be represented as ${\rm NGEO}(s, s^*) = min(\frac{{\rm GEO}(s, s^*)}{ \vert s^* \vert }, 1)$ where $\vert s^* \vert $ is the length of ground truth sequence. ${\rm GEO}(s, s^*)$ is the number of operations required to convert $s$ to $s^*$, which determine semantic similarity between two DBPedia entities/relations based on DBPedia's ontology structure. As a result, lower NGEO suggests better path correctness.  		
		For \textit{path well-formedness}, we report the average \textbf{\% of ill-formatted paths (\%IF)} and \textbf{\% of invalid relations (\%IV)}. An ill-formatted path is one that does not alternate between entities and relations in its  sequence, or at least one of its entities and relations does not follow the DBPedia entity and ontology property format, i.e., \textsf{dbr:ENTITY - \textit{dbo:RELATION} - ... - \textit{dbo:RELATION} - dbr:ENTITY}. %As we want the model to generate a specific output format, we do not conduct any post-processing on the output and 
		We also consider the path ill-formatted if the LLM generates multiple one-hop relations instead of a multi-hop path. The \%IV measure is defined by the proportion of invalid relations in the generated path. We define a relation to be invalid if it involves ontology hallucination. 
		We conduct five rounds of generation over all CPG tasks and report the averages of the above metrics across all the generated paths.
		
		\subsubsection{Models}
		In this experiment, we employed the same three text-based large language models as previous experiments under the same parameter setting. Please refer to Appendix~\ref{app:prompt_design} for the prompt used for each of the LLMs.
		Additionally, we include two simple baseline methods:
		
		\noindent\textbf{The Shortest Path baseline} returns the shortest path between the query entities in the knowledge graph downloaded from Dbpedia. Note that this method does not involve any LLM, and always return well-formed paths. 
		
		\noindent\textbf{The Simple Instruction  baseline} prompts a text-based LLM with the context document followed by the instruction below: \textit{``Instruction: Generate the contextual path between $e_H$ and $e_T$.''} The prompt does not explain to LLM what a contextual path is.  
		
		All experiments are conducted under zero-shot setting.
	
		\subsubsection{Multi-Step Prompting}
		Unlike the earlier knowledge path generation task, CPG involves an additional context document, and requires the generation of knowledge path to be relevant to the context. 
		Earlier LLMs such as \textsf{text-davinci-003} might struggle with solving such complex tasks, as suggested in previous works~\cite{Khot_Trivedi_Finlayson_Fu_Richardson_Clark_Sabharwal_2022}. As a result, we decompose CPG into small and simple subtasks, which is an effective strategy for several other complex tasks~\cite{,Katz_Geva_Berant_2022,Fu_Peng_Sabharwal_Clark_Khot_2022,Creswell_Shanahan_Higgins_2022}.
		Our proposed multi-step prompting strategy performs CPG with the following subtasks: 
		
		\noindent\textbf{Support Sentence Extraction.}
		We design this subtask to prompt LLM to find query-relevant \textit{support sentences} in the context document, as previous works have shown that LLMs can be easily distracted by irrelevant input~\cite{Shi_Chen_Misra_Scales_Dohan_Zhou_2023}.
		The support sentences are then passed to LLMs during path generation. They provide concise information about the relationship between query entities in the context document, which prevents Codex to generate non-factual relations and thus reduce content hallucination. 
		
		\noindent\textbf{Entity Linking.}
		With a set of support sentences, this subtask links the query entities in the sentences to their corresponding DBPedia entities in the format of \textsf{dbr:ENTITY\_NAME}. 
		
		\noindent\textbf{Path Generation.}
		Finally, this subtask generates the contextual path based on the support sentences and the linked query entities.
		See Appendix~\ref{app:prompt_design} for prompt designed based on multi-step prompting strategy.
		
		\begin{table}[t!]
			\small
			\caption{Contextual Path Generation}
			\label{tbl:CPG}
			\begin{tabular}{llccc}
				\toprule
				&  & \multicolumn{1}{l}{\textbf{NGEO}} & \multicolumn{1}{l}{\textbf{\%IF}} & \multicolumn{1}{l}{\textbf{\%IV}} \\
				\midrule
				\multirow{2}{*}{\begin{tabular}[c]{@{}l@{}}Simple\\ Baseline\end{tabular}} & Shortest Path & 0.43 & - & - \\
				& Simple Instruction & 0.44 & 0.62 & 0.37 \\
				\midrule
				\multirow{2}{*}{Text-Davinci-003} & Single-Step & 0.37 & 0.46 & 0.34 \\
				& Multi-Step & 0.29 & 0.35 & 0.3 \\
				\midrule
				ChatGPT & Single-Step & 0.29 & 0.15 & 0.31 \\
				& Single-Step + AutoCoT & 0.25 & 0.14 & 0.27 \\
				\midrule
				GPT-4 & Single-Step & \textbf{0.17} & 0.1 & \textbf{0.18}\\
				\bottomrule
			\end{tabular}
			\vspace{-4mm}
		\end{table}

	\subsubsection{Result}
	The results are presented in Table~\ref{tbl:CPG}. \textsf{GPT-4} outperforms its predecessors by a significant margin due to three main reasons: (1) it generates concise outputs, (2) it exhibits lower hallucination rates compared to other models, and (3) it follows instructions more accurately, resulting in higher well-formedness. It is also worth noting that \textsf{GPT-4} almost does not require any prompt or answer engineering. Remarkably, \textsf{ChatGPT} also underperforms in the Single-Step Setting, due to poor well-formedness of the generated pass. However, its performance improves when additional efforts are introduced such as AutoCoT, leading to enhanced well-formedness and correctness. Still, one major drawback of \textsf{ChatGPT} is lack of conciseness. It often generate a lot of noises and require further answer engineering.
	Finally, \textsf{text-davinci-003} requires multi-step prompting to generate correct and well-formed outputs. Despite introducing multi-step prompting, we do not observe a significant improvement in \%IV. This suggests that more effort is required in prompt engineering to utilize \textsf{text-davinci-003} for complex knowledge graph reasoning problems.

	As the task of CPG is inherently challenging, we have observed a significant increase in the incidence of both types of hallucinations. We acknowledge that reducing hallucinations in KG reasoning tasks remains a challenge that requires further investigation in future research.

	\section{Conclusion}
	In this work, we propose two research questions which explore the potential to leverage Large Language Models' internal knowledge graph to address knowledge graph reasoning tasks. The result suggests 
	As a preliminary study, the key findings in this work provide great potential for future research. Our prompt design can be further extended to address reasoning over proprietary knowledge graphs which may not have been seen by LLMs during their training, as well as to handle other knowledge graph-based recommendation and information retrieval problems.
	
	\bibliographystyle{ACM-Reference-Format}
	\bibliography{0_Ref}
	
	\appendix
	
	\begin{table*}[t!]
		\small
		\caption{Examples: Non-contextual Relation Generation}
		\label{tbl:exp_nc_rel_gen}
		\begin{tabular}{cccc}
			\toprule
			\multicolumn{1}{c}{\textbf{Ground Truth}} & \multicolumn{1}{c}{\textbf{text-davinci-003}} & \multicolumn{1}{c}{\textbf{ChatGPT}} & \multicolumn{1}{c}{\textbf{GPT-4}} \\
			\midrule
			\multicolumn{4}{l}{\textbf{QUERY:} dbr:Tomb\_Raider\_(soundtrack) - dbp:label - ?} \\
			\midrule
			dbr:Sony\_Classical\_Records & \begin{tabular}[c]{@{}c@{}}dbr:Tomb\_Raider\_\\ (Original\_Motion\_Picture\_Soundtrack)\end{tabular} & \begin{tabular}[c]{@{}c@{}}dbr:Tomb\_Raider:\_\\ Original\_Motion\_Picture\_Soundtrack\end{tabular} & dbr:Milan\_Record \\
			\midrule
			\multicolumn{4}{l}{\textbf{QUERY:} dbr:Tim\_Burton - dbo:education - ?} \\
			\midrule
			\begin{tabular}[c]{@{}c@{}}dbr:Burbank\_High\_School\_\\ (Burbank,\_California)\end{tabular} & dbr:California\_Institute\_of\_the\_Art & dbr:California\_Institute\_of\_the\_Art & dbr:California\_Institute\_of\_the\_Art
			\\ \bottomrule
		\end{tabular}
	\end{table*}
	
	\begin{table*}[]
		\small
		\caption{Example: Contextual Relation Generation}
		\label{tbl:c_rel_gen}
		\begin{tabular}{ll}
			\toprule
			\multicolumn{2}{c}{\textbf{Example 1}}\\
			\midrule
			
			\textbf{Head Entity} & dbr:Playtone \\
			\textbf{Tail Entity} & dbr:Tom\_Hanks \\
			\multicolumn{2}{l}{\textbf{Context}} \\
			\multicolumn{2}{l}{\begin{tabular}[c]{@{}l@{}}\underline{Playtone} is an American film and television production company\\ established in 1998 by actor \underline{Tom Hanks} and producer Gary Goetzman.\end{tabular}} \\
			\midrule
			
			\multicolumn{2}{l}{\textbf{Ground Truth}} \\
			\multicolumn{2}{l}{dbr:Playtone - dbo:founder - dbr:Tom\_Hanks} \\
			\midrule

			\multicolumn{2}{l}{\textbf{Text-davinci-003}} \\
			\multicolumn{2}{l}{dbr:Playtone, dbo:founder, dbr:Tom\_Hanks} \\
			\midrule

			\multicolumn{2}{l}{\textbf{ChatGPT}} \\
			\multicolumn{2}{l}{dbr:Playtone - dbo:owner - dbr:Tom\_Hanks} \\
			\midrule

			\multicolumn{2}{l}{\textbf{GPT-4}} \\
			\multicolumn{2}{l}{dbr:Playtone, dbo:founder, dbr:Tom\_Hanks}
			\\ \midrule

			\multicolumn{2}{c}{\textbf{Example 2}}\\
			\midrule
			\textbf{Head Entity} & dbr:The\_Big\_Short\_(film) \\
			\textbf{Tail Entity} & dbr:Brad\_Pitt \\
			\multicolumn{2}{l}{\textbf{Context}} \\
			\multicolumn{2}{l}{\begin{tabular}[c]{@{}l@{}}\underline{The Big Short} is a 2015 American biographical crime comedy-drama\\ film directed and co-written by Adam McKay. On January 13, 2015,\\Variety reported that Christian Bale, and Ryan Gosling were set to\\star in the film, with \underline{Pitt} producing the film along with Dede Gardner\\and Jeremy Kleiner. 
			\end{tabular}} \\
			\midrule
			
			\multicolumn{2}{l}{\textbf{Ground Truth}} \\
			\multicolumn{2}{l}{dbr:The\_Big\_Short\_(film) - dbo:producer - dbr:Brad\_Pitt} \\
			\midrule

			\multicolumn{2}{l}{\textbf{Text-davinci-003}} \\
			\multicolumn{2}{l}{dbr:The\_Big\_Short\_(film) - dbo:starring - dbr:Brad\_Pitt} \\
			\midrule

			\multicolumn{2}{l}{\textbf{ChatGPT}} \\
			\multicolumn{2}{l}{dbr:The\_Big\_Short\_(film) - dbo:producer - dbr:Brad\_Pitt} \\
			\midrule

			\multicolumn{2}{l}{\textbf{GPT-4}} \\
			\multicolumn{2}{l}{dbr:The\_Big\_Short\_(film) - dbo:producer - dbr:Brad\_Pitt}
			\\ \bottomrule
			
		\end{tabular}
	\end{table*}

	\section{Contextual Path Generation Tasks}
		\subsection{Prompt Design}\label{app:prompt_design}
			\subsubsection{Text-Davinci-003} 
   \hphantom\\

			\textbf{Subtask 1: Support Sentence Extraction} 
			\begin{displayquote}
				Instruction: Find the sentences from the following context document that explain the relationship between \textbf{<head entity>} and \textbf{<tail entity>}.\\
				Context Document: \textbf{<context document>}
			\end{displayquote}
			
			\textbf{Subtask 2: Entity Linking}
			\begin{displayquote} 
				According to the following context, link both \textbf{<head entity>} and \textbf{<tail entity>} to their corresponding DBpedia entities in the format of dbr:ENTITY\_NAME.\\
				Context: \textbf{<support sentences>}
			\end{displayquote}
			
			\textbf{Subtask 3: Path Generation}
			\begin{displayquote}
				Insert DBPedia entities and ontology properties between \textbf{<head entity>} and \textbf{<tail entity>} to form one single multi-hop path which reflects the relationship between them in the following context. Pay attention to the path format dbr:ENTITY\_NAME, dbo:ONTOLOGY, dbr:ENTITY\_NAME, ..., dbo:ONTOLOGY, dbr:ENTITY\_NAME\\
				Context: \textbf{<support sentences>}
			\end{displayquote}
		
			\subsubsection{ChatGPT}
			\begin{displayquote}
				Context: \textbf{<context document>}\\
				Instruction: Insert DBPedia entities and ontology properties between \textbf{<head entity>} and \textbf{<tail entity>} to form one single multi-hop path which reflects the relationship between them in the context. Pay attention to the path format dbr:ENTITY\_NAME, dbo:ONTOLOGY, dbr:ENTITY\_NAME, ..., dbo:ONTOLOGY, dbr:ENTITY\_NAME
			\end{displayquote}
			
			\textbf{AutoCoT} 
			\begin{displayquote}
				Let's think step by step.
			\end{displayquote}
			
			\subsubsection{GPT-4}
			\begin{displayquote}
				Context: \textbf{<context document>}\\
				Instruction: Insert DBPedia entities and ontology properties between \textbf{<head entity>} and \textbf{<tail entity>} to form one single multi-hop path which reflects the relationship between them in the context. Pay attention to the path format dbr:ENTITY\_NAME, dbo:ONTOLOGY, dbr:ENTITY\_NAME, ..., dbo:ONTOLOGY, dbr:ENTITY\_NAME
			\end{displayquote}

\end{document}